\documentclass[11pt,a4paper]{article}
\usepackage[hyperref]{emnlp2020}
\usepackage{times}
\usepackage{latexsym}

\usepackage{microtype}

\aclfinalcopy % Uncomment this line for the final submission
\usepackage{times}
\usepackage{latexsym}
\usepackage{indentfirst}
\usepackage{graphicx}
\usepackage[ruled,vlined]{algorithm2e}
\usepackage{amsmath}
\usepackage{mathtools}
\usepackage{footnote}

\usepackage[T1]{fontenc}
\usepackage[utf8]{inputenc}

\usepackage{microtype}

\title{Multilingual Code-Switching for Zero-Shot Cross-Lingual \\ Intent Prediction and Slot Filling}

\author{Jitin Krishnan \quad Antonios Anastasopoulos \quad Hemant Purohit \quad Huzefa Rangwala \\ George Mason University \\ Fairfax, VA, USA \\ 
\texttt{\{jkrishn2,antonis,hpurohit,rangwala\}@gmu.edu}
}

\begin{document}
\maketitle
\begin{abstract}
Predicting user intent and detecting the corresponding slots from text are two key problems in Natural Language Understanding (NLU). 
In the context of zero-shot learning, this task is typically approached by either using representations from pre-trained multilingual transformers such as mBERT, or by machine translating the source data into the known target language and then fine-tuning. 
Our work focuses on a particular scenario where the target language is \textit{unknown} during training. To this goal, we  %show that augmenting
propose a novel method to augment the monolingual source data using multilingual code-switching via random translations 
%and show that it can 
to enhance a transformer's language neutrality when fine-tuning it for a downstream task. 
% 
%Along with producing a 
%We also present
This method also helps discover novel insights %show how our method can help in shedding interesting linguistic
on how code-switching with different language families around the world %and studying their
impact the performance on the target language. 
%and  into how different language families around the world impact individual languages. 
% 
Experiments on the benchmark dataset of MultiATIS++ yielded an average improvement of $+4.2\%$ in accuracy for intent task and $+1.8\%$ in F1 for slot task using our method over the state-of-the-art across $8$ different languages\footnote{Languages that have different morphological structures compared to English, such as Hindi, Turkish, Chinese, and Japanese, yielded higher benefits.}.  
Furthermore, we present an application of our method for crisis informatics using a new human-annotated tweet dataset of slot filling in English and Haitian Creole, collected during Haiti earthquake disaster\footnote{Dataset and implementation available at \url{https://github.com/jitinkrishnan/Multilingual-ZeroShot-SlotFilling}.}. 

\end{abstract}

%t-SNE plot of embeddings retrieved from the transformer blocks of multilingual BERT for English (black), Chinese (cyan), French (blue), German (green), and Japanese (red) sentences in the MutiATIS++ dataset.
\begin{figure*} [hbt!]
    \centering
    \includegraphics[scale=0.277]{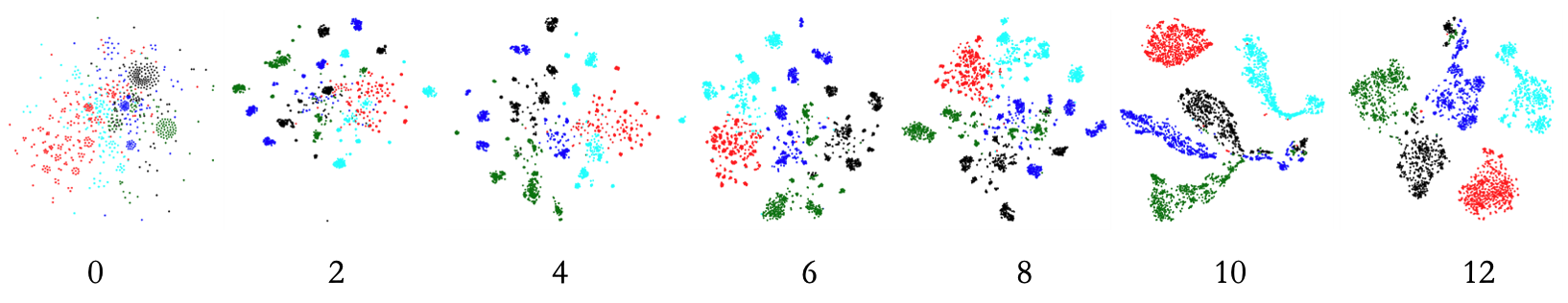}
    \caption{t-SNE plot of embeddings across the $12$ multi-head attention layers of multilingual BERT. Parallelly translated sentences of MutiATIS++ dataset are still clustered according to the languages: English (black), Chinese (cyan), French (blue), German (green), and Japanese (red).} \label{fig:mbert}
\end{figure*}

\begin{figure*} [ht]
    \centering
    \includegraphics[scale=0.255]{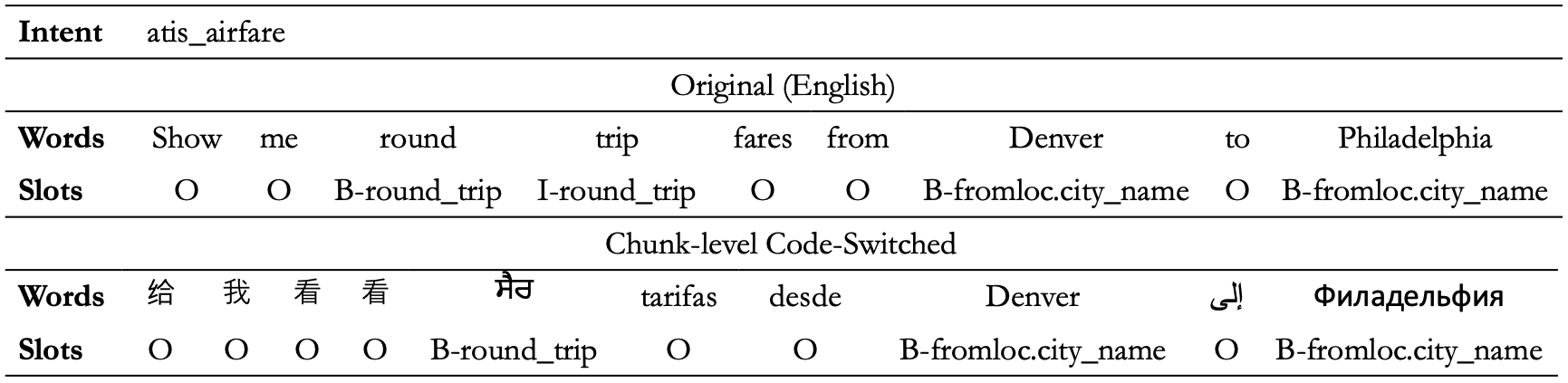}
    \caption{An original example in English from MultiATIS++ dataset and its multilingually code-switched version. In the above code-switching example, the chunks are in Chinese, Punjabi, Spanish, English, Arabic, and Russian. `\textit{atis\_airfare}' represents an intent class where the user %mentions
    seeks price of a ticket.} \label{fig:chunk}
\end{figure*}
% Even when the translations are not entirely accurate, in our experiments, this did not reduce the performance when augmented to the translate-train model.

% The language clusters are still visible in a pre-trained model.

\section{Introduction}
\noindent A cross-lingual setting is typically described as a scenario in which a model trained for a particular task in one language (e.g. English) should be able to generalize well to a different language (e.g. Japanese). While a semi-supervised solution \cite{xiao2013semi,muis2018low} assumes some target language data is available, a zero-shot solution \cite{eriguchi2018zero,srivastava2018zero,xu2020end} assumes none is available at training time. This is particularly significant in real world  problems such as extracting relevant information during a new disaster   \cite{nguyen2017robust,krishnan2020unsupervised} and hate speech detection \cite{pamungkas2019cross,stappen2020cross}, where the target language might be of low-resource or unknown. In such scenarios, it is crucial that models can generalize well to unseen languages. 

Intent prediction and slot filling are important NLU tasks and significant for real world problems. % such as those discussed above. 
They are studied extensively for goal-oriented dialogue systems currently, such as Amazon's Alexa, Apple's Siri, Google Assistant, and Microsoft's Cortana. Finding the `intent' behind the user's query and identifying relevant `slots' in the sentence to engage in a dialogue are essential for an effective conversational assistance. For example, users might want to `\textit{play music}' given the slot labels `\textit{year}' and `\textit{artist}' \cite{coucke2018snips}, or they may want to `\textit{book a flight}' given the slot labels `\textit{airport}' and `\textit{locations}' \cite{price1990evaluation}. A strong correlation between the two tasks has made jointly trained models successful \cite{goo2018slot, niu2019novel, hardalov2020enriched, chen2019bert}. In a cross-lingual setting, the model should be able to learn this joint task in one language and transfer knowledge to another \cite{upadhyay2018almost,schuster2018cross,xu2020end}. This is the premise of our work.

Highly effective multilingual models such as mBERT \cite{devlin2018bert} and XLM-R \cite{conneau2020unsupervised} have shown success across several multilingual tasks in recent years. 
In the zero-shot cross-lingual transfer setting with an unknown target language, 
a typical solution is to use pre-trained transformer models and fine-tune to the downstream task using the monolingual source data \cite{xu2020end}. 
However, previous work \cite{pires2019multilingual} has shown that existing transformer-based representations may exhibit systematic deficiencies for certain language pairs.Previous work \cite{pires2019multilingual} has shown that existing transformer-based representations may exhibit systematic deficiencies for certain language pairs. Figure \ref{fig:mbert} shows that the representations across the $12$ multi-head attention layers of mBERT are still clustered according to the languages.
This leads to a fundamental challenge that we address in this work: enhancing the language neutrality so that the fine-tuned model is generalizable across languages for the downstream task. 
To this goal, we introduce a data augmentation method via multilingual code-switching, where the original sentence in English is code-switched into randomly selected languages. For example, chunk-level code-switching creates sentences with phrases in multiple languages as shown in Figure \ref{fig:chunk}. 
We show that this can lead to a better performance in the zero-shot %cross-lingual transfer 
setting such that mBERT can be fine-tuned for all languages (not just one) with a monolingual source data. 

Further, we show how code-switching with different language families impact the model's performance on individual target languages. 
Cross-lingual study of language families largely remains unexplored for NLU tasks. % of intent and slots. 
For instance, while it might be intuitive that Sino-Tibetan language family can aid a task in Hindi, results indicating that Turkic language family may help Japanese can reveal intriguing inter-family relationships and how they are aligned in the underlying language model's vector space. % that can foster further linguistic research and aid low resource languages.

\textbf{Contributions}: 
\textbf{a)} We present a data augmentation method via multilingual code-switching to enhance the language neutrality of mBERT for fine-tuning to a downstream NLU task of intent prediction and slot filling. 
\textbf{b)} By code-switching into different language families, we show that potential relationships between a family and a target language can be identified and studied; which could help foster zero-shot cross-lingual research in low-resource languages.  
\textbf{c)} We release a new human-annotated tweet dataset, collected during Haiti earthquake disaster, for intent prediction and slot filling in English and Haitian Creole. 

\textbf{Advantages}: With enhanced generalizability, our model can be deployed with an out-of-the-box functionality. Previous methods of first machine translation of the source data into the known target language, followed by fine-tuning (referred `translate-train') \cite{xu2020end, yarowsky2001inducing, shah2010synergy, ni2017weakly} require a separate model to be trained for each language. % is seen during test time. 
%A key advantage of our work is that only one fine-tuned model is needed for the downstream task for all languages.

\section{Related Work}

\subsection{Cross-Lingual Transfer Learning}

\noindent Researchers have studied cross-lingual tasks in various settings such as sentiment/sequence classification \cite{wan2009co,eriguchi2018zero,yu2018multilingual}, named entity recognition \cite{zirikly2015cross,tsai2016cross,xie2018neural}, parts-of-speech tagging \cite{yarowsky2001inducing,tackstrom2013token,plank2018distant}, and natural language understanding \cite{he2013multi,upadhyay2018almost,xu2020end}. The methodology for most of the current approaches for cross-lingual tasks fall into the following three categories: \textbf{a)} multilingual representations from pre-trained or fine-tuned models such as mBERT \cite{devlin2018bert} or XLM-R \cite{conneau2020unsupervised}, \textbf{b)} machine translation followed by alignment \cite{shah2010synergy, yarowsky2001inducing, ni2017weakly}, or \textbf{c)} a combination of both \cite{xu2020end}. Before %the arrival of 
transformer models, effective approaches included domain adversarial training to extract language-agnostic features \cite{ganin2016domain,chen2018adversarial} and word alignment methods such as MUSE \cite{conneau2017word} to align fastText word vectors \cite{bojanowski2017enriching}. Recently, \citeauthor{conneau-etal-2020-emerging}, \citeyear{conneau-etal-2020-emerging} has shown that having shared parameters in the top layers of the multi-lingual encoders can be used to align different languages quite effectively on tasks such as XNLI \cite{conneau2018xnli}.

Monolingual models for joint slot filling and intent prediction have used methods such as attention-based RNN \cite{liu2016attention} and attention-based BiLSTM with a slot gate \cite{goo2018slot} on benchmark datasets such as ATIS \cite{price1990evaluation} and SNIPS \cite{coucke2018snips}. These methods have shown that a joint method can enhance both tasks and slot filling can be conditioned on the learned intent. An interrelated mechanism was introduced \cite{niu2019novel} to iteratively learn the relationship between the two tasks. Recently, BERT-based approaches \cite{hardalov2020enriched,chen2019bert} have shown improved results. On the other hand, cross-lingual versions of this joint task include a low-supervision based approach for Hindi and Turkish \cite{upadhyay2018almost}, new dataset for Spanish and Thai \cite{schuster2018cross}, and the most recent work of MultiATIS++ \cite{xu2020end} creating a comprehensive dataset in 9 languages; which is used to benchmark our results.

%\subsection{Zero-Shot Learning}

%\noindent 
The joint task mentioned above in a pure zero-shot learning is the motivation of our work. Zero-shot is described as the setting where the model sees a new distribution of examples during test time \cite{xian2017zero,srivastava2018zero,romera2015embarrassingly}. It is common for machine translation based methods to translate source data to the target language before training. We assume that target language is unknown during training, so that our model is generalizable across languages. 

%the model we build can be used out-of-the. This way, no model training is required when new data is coming in. i.e., the model will be target language agnostic.

\subsection{Code-Switching}
\noindent Linguistic code-switching is a phenomenon where multilingual speakers alternate between languages. 
%It is a growing research direction, but still remains a challenge because code-switching usually occurs during speech or chat. 
Recently, monolingual models have been adapted to code-switched text in several tasks such as entity recognition \cite{aguilar2019english}, part-of-speech tagging \cite{soto2018joint,ball2018part}, sentiment analysis \cite{joshi2016towards}, and language identification \cite{mave2018language,yirmibecsouglu2018detecting,mager2019subword}. Recently, \citeauthor{khudabukhsh2020harnessing}, \citeyear{khudabukhsh2020harnessing} have proposed a pipeline to sample code-mixed documents using minimal supervision. \citeauthor{qin2020cosda}, \citeyear{qin2020cosda} allows randomized code-switching to include the target language. In our context, if the target language is German, we ensure that there is no code-switching to German during training. We consider this distinction essential to evaluate a true zero-shot learning scenario and prevent any bias. Another recent work by \citeauthor{yang2020alternating}, \citeyear{yang2020alternating} presents a non-zero-shot approach that performs code-switching to target languages. \citeauthor{jiang2020multilingual}, \citeyear{jiang2020multilingual} presents presents a code-switching based method to improve the ability of multilingual language models for factual knowledge retrieval. Code-switching is usually done at the word-level. However, our results favor chunk-level switching over word-level as the latter may bring more noise to the code-switched version when compared to the original meaning of the sentence. Code-switching and other data augmentation techniques have been applied to the pre-training stage in recent works \cite{chaudhary2020dict,dufter2020identifying}, however we do not address pre-training in this work. Pre-trained models such as XLM-R is also likely to be exposed to code-switched data, as it is trained using common-crawl. In this work, we specifically focus on mBERT which largely remain monolingual at the sentence level to identify the impact of code-switching during fine-tuning, in addition to study the impact of language-family-based augmentations.

%. i.e., we take a monolingual sentence and randomly code-switch them at word-level, chunk-level, or sentence-level. 

%Exploring code-switching to improve the masked language model objective is left as future work.

%We show this augmentation helps boost performance of 

%Note that the amount of code-switched data present in the massive multilingual corpora used for pre-training transformer models such as BERT is low because Wikipedia rarely contain intra-sentence code-switching. We exploit this gap to enhance the language neutrality of the transformer by augmenting it with artificially code-switched data.

\section{Methodology}
This section first describes our problem for zero-shot cross-lingual transfer setting, followed by a novel data augmentation method using multilingual code-switching of monolingual source to enhance language neutrality. We then describe language families, followed by the joint training setup. 
%as well as the language families to deeper analysis. 
% 
\subsection{Problem Definition}

\iffalse
\begin{table}[!htbp]
%\footnotesize
%\scriptsize
\small
\begin{center}
 \begin{tabular}{p{1.0cm} p{5.6cm} } 
 \hline
 \textbf{Notation} & \textbf{Definition} \\ [0.5ex] 
 \hline
 $X_{ut}^L$ & Utterances in language $L$ \\
 $y_{i}^L$ & Ground truth intent classes for the Utterances $X_{ut}^L$ \\
 $y_{sl}^L$ & Ground truth slot labels for the Utterances $X_{ut}^L$ \\
 $l_{T}$ & Target Language (Test Language)  \\
 $sty$ & Represents Code-Switching \\
 $|d|$ & Source dataset size \\
 $k$ & Number of code-switchings per sentence \\
 $\alpha$ & Joint training weight on Intent \\
 $\beta$ & Joint training weight on Slot \\
 \hline
 \end{tabular}
\end{center}
\caption{Notations}\label{tab:notations}
\end{table}
\fi

\noindent Given a source (S) and a set of target (T) languages, the goal is to train a classifier using data only in the source language and predict examples from the completely unseen target languages.   %Additionally, we add the 
We assume the target language is \textit{unknown} during training time, which makes direct translation to target infeasible. In this context, we use code-switching ($cs$) to augment the monolingual source data. Thus, the input and output of our problem can be defined as: \\
\textbf{\textit{Input:}}  $X_{ut}^S$, $y_{i}^{S}$, $y_{sl}^{S}$\\
\textbf{\textit{Code-Switched Input:}} $X_{ut}^{cs}$, $y_{i}^{cs}$, $y_{sl}^{cs}$ \\
\textbf{\textit{Output:}} $y_{i}^{T}$, $y_{sl}^{T}$ $\gets predict(X_{ut}^{T})$\\
\text{\ \ \ } where $X_{ut}$ represents sentences, $y_{i}$ their ground truth intent classes, and  $y_{sl}$ the slot labels for the words in those sentences. An example sentence, its intent class, and slot labels are shown in Figure  \ref{fig:chunk}. 

%Code-switched data is represented as: $X_{ut}^{cs}$, $y_{i}^{cs}$, $y_{sl}^{cs}$.  where Xut is utterances or sentences, yi is the 212 ground truth intent classes and ysl slot labels for 213 those utterances.

%And, target language is represented as $l_{T}$

\subsection{Multilingual Code-Switching}
% The goal of mixing other languages in the same sentence or in a monolingual dataset is to push the language-neutrality of the model. This can help generalize better when introduced to a new language.
%Multilingual masked language models, such as mBERT \cite{devlin2018bert}, that are trained using large datasets of publicly available unlabeled corpora such as Wikipedia remain largely monolingual at the sentence level.

\noindent Multilingual masked language models, such as mBERT \cite{devlin2018bert}, are trained using large datasets of publicly available unlabeled corpora such as Wikipedia.  Such corpora largely remain monolingual at the sentence level because the presence of intra-sentence code-switched data in written texts is likely scarce. 
%Code-switched data present in such massive multilingual corpora is likely scarce because Wikipedia rarely contain intra-sentence code-switching.
%We exploit this gap to enhance the language neutrality of the transformer by augmenting it with artificially code-switched data.
The masked words that needed to be predicted usually are in the same language as their surrounding words. 
%We aim to fill this gap via multilingual code-switching procedure which can enhance the language-neutrality of the pre-trained models while fine-tuning to a downstream task.
We study how code-switching can enhance the language neutrality of such language models by augmenting it with artificially code-switched data for fine-tuning it to a downstream task.
Algorithm \ref{alg:style} explains this code-switching process at the chunk-level. When using slot filling datasets, slot labels that are grouped by BIO \cite{ramshaw1999text} tags constitute natural chunks. To summarize the algorithm, we take a sentence, take each chunk from that sentence, perform a translation into a random language using Google's NMT system \cite{wu2016google}, and align the slot labels to fit the translation. At the chunk-level, we use a direct alignment. i.e., the BIO-tagged labels are recreated for the translated phrase based on the word tokens.  More complex methods can be applied here to improve the alignment of the slot labels such as fast-align \cite{dyer2013simple} or soft-align \cite{xu2020end}. Code-Switching at the word-level essentially translates every word randomly, while at the sentence-level translates the entire sentence. During the experimental evaluation process, to build a language neutral model using monolingual source of English data, \textbf{all $8$ target languages are excluded} from the code-switching procedure to avoid unfair model comparisons,  
%prevent any performance skews,
i.e. remove target languages from % $googletrans.languages  (
$lset$ in Algorithm \ref{alg:style}.

%- target.languages$.
%($l_{T}$)

\noindent \textit{Complexity}: The augmentation process is repeated $k$ times per sentence producing a new augmented dataset of size $k\times n$, where $n$ is the size of the original dataset, i.e. space complexity of $\mathcal{O}(k\times n)$. Algorithm \ref{alg:style} %procedure 
has a runtime complexity of $\mathcal{O}(k\times n \times translations / sentence)$ steps assuming constant time for alignment.  Word-level requires as many translations as the number of words but sentence-level requires only one. An increase in the dataset size also increases the training time, but an advantage is %that 
one model fits all languages.

\begin{algorithm}[t]
\SetAlgoLined
\textbf{Input}: $X_{ut}^{en}$, \  $y_i^{en}$, \ $y_{sl}^{en}$\\
\textbf{Output}: $X_{ut}^{cs}$, \  $y_{i}^{cs}$, \  $y_{sl}^{cs}$ \\
$X_{ut}^{cs} \gets \emptyset $, \ $y_{i}^{cs} \gets \emptyset $, \ $y_{sl}^{cs} \gets \emptyset $ \\
$lset = googletrans.languages - l_{T}$ \\
\For{$i \in 1.. \ k$}{
    \For{$i \in 1..\ len(X_{ut}^{en}) $}{
        $G^{cs} \gets \emptyset $, $L^{cs} \gets \emptyset $ \\
        $chunks = slot\_chunks(X_{ut}^{en}[i], \ y_{sl}^{en}[i])$ \\
        \For{$c \in chunks$}{
            $l \gets random.choice(lset)$ \\
            $t \gets translate(c,l)$ \\
            $G^{cs} \gets G^{cs} \cup t$ \\
            $L^{cs} \gets L^{cs} \cup align\_label(c, t)$ \\
        }
        $X_{ut}^{cs} \gets X_{ut}^{cs} \ \cup \ G^{cs}$ \\ 
        $y_{i}^{cs} \gets y_{i}^{cs} \ \cup \ y_i^{cs}[i]$ \\ 
        $y_{sl}^{cs} \gets y_{sl}^{cs} \ \cup \ L^{cs}$ \\
    }
}
 \caption{Data Augmentation via Multilingual Code-Switching (Chunk-Level)} \label{alg:style}
\end{algorithm}

\begin{table*}
%\resizebox{\columnwidth}{!}{%
%\small
%\footnotesize
\centering
\scalebox{0.9}{
\begin{tabular}{cc}
    \hline
 \textbf{Group Name} & \textbf{Languages} \\ 
 \hline
 Afro-Asiatic & Arabic (ar), Amharic (am), Hebrew (he), Somali (so) \\ 
 Germanic & German (de), Dutch (nl), Danish (da), Swedish (sv), Norwegian (no)  \\
 Indo-Aryan & Hindi (hi), Bengali (bn), Marathi (mr), Nepali (ne), Gujarati (gu), Punjabi (pa) \\ 
 Romance & Spanish (es), Portuguese (pt), French (fr), Italian (it), Romanian (ro)  \\
 Sino-Tibetan \& Japonic & Chinese (zh-cn), Japanese (ja), Korean (ko)\\ %Mongolian (mn), Turkish (tr) \\ 
 Turkic & Turkish (tr), Azerbaijani (az), Uyghur (ug), Kazakh (kk)  \\
    \hline
\end{tabular}
}
 \caption{Selected language families to evaluate their impact on a target language.} \label{table:lgroups}
\end{table*}

\begin{table*}
%\scriptsize
%\footnotesize
%\small
\begin{center}
\begin{tabular}{c*{3}{c}*{3}{c}*{2}{c}}
\hline
Language    & \multicolumn{3}{c}{Utterances} & \multicolumn{3}{c}{Tokens}  & Intents & Slots       \\
            & train & dev & test & train & dev & test & & \\
\hline
\multicolumn{9}{c}{MultiATIS++ \cite{xu2020end}} \\
\hline
English   & 4488   & 490 & 893 & 50755 & 5445 & 9164 & 18 & 84   \\
Spanish  & 4488 & 490 & 893 & 55197 & 5927 & 10338 & 18 & 84 \\
%  &  &  &  & \textbf{. . .}  &  &  &  &  \\
Portuguese  & 4488 & 490 & 893 & 55052 & 5909 & 10228 & 18 & 84 \\
German  & 4488 & 490 & 893 & 51111 & 5517 & 9383 & 18 & 84 \\
French  & 4488 & 490 & 893 & 55909 & 5769 & 10511 & 18 & 84 \\
Chinese  & 4488 & 490 & 893 & 88194 & 9652 & 16710 & 18 & 84 \\
Japanese  & 4488 & 490 & 893 & 133890 & 14416 & 25939 & 18 & 84 \\
Hindi  & 1440 & 160 & 893 & 16422 & 1753 & 9755 & 17 & 75 \\
Turkish  & 578 & 60 & 715 & 6132 & 686 & 7683 & 17 & 71 \\
\hline
\multicolumn{9}{c}{Disaster Tweets (New Dataset)} \\
    \hline
English   & 3518  & 490  & - & 16369 & 4242 & - & 2 & 5   \\
Haitian Creole  & - & - & 520 & - & - & 2834 & 2 & 5  \\
\hline
\end{tabular}
\end{center}
 \caption{Datasets and statistics.} \label{table:stat}
\end{table*}

\iffalse
\begin{figure} [ht]
    \centering
    \includegraphics[scale=0.25]{joint3.png} %[width=15cm, height=8cm]
    \caption{Settings for joint training using mBERT \citeauthor{devlin2018bert}.} \label{fig:joint}
\end{figure}
\fi

\subsection{Language Families}

\noindent A language family is defined as a group of related languages that are likely coming from the same parent. For example, Portuguese, Spanish, French, Italian, and Romanian are daughter languages derived from Latin \cite{languagefamily}. We use language families to study their impact on the target languages. %This is done by 
We augment %ing 
the source language with code-switching from a particular language family. For instance, code-switching the English dataset with Turkic language family and testing on Japanese can reveal %their linguistic connection or 
how closely the two are aligned in the vector space of a pre-trained multilingual model. From a set of 5 distinct language families, we select a total of 6 groups of languages: Afro-Asiatic \cite{voegelin1976classification}, Germanic \cite{harbert2006germanic}, Indo-Aryan \cite{masica1993indo}, Romance \cite{elcock1960romance}, Sino-Tibetan and Japonic \cite{shafer1955classification, miller1967japanese}, and Turkic \cite{johanson2015turkic}. Germanic, Romance, and Indo-Aryan are branches of the Indo-European language family. Language groups and their selected daughter languages are shown in Table \ref{table:lgroups}. Each group is selected based on a target language in the dataset and Afro-Asiatic family is added as an extra group. %For the 
In experiments, %$googletrans.languages$ 
$lset$ in Algorithm \ref{alg:style} will be %replaced by
assigned languages from a specific family. % $familyName.languages$. 

%a To summarize, each family of code-switched data is augmented to the original English data and trained to evaluate on the $8$ target languages.

\begin{table*}
%\resizebox{\columnwidth}{!}{%
\centering
\scalebox{0.79}{
\begin{tabular}{l|c|cccccccc|c}
    \hline
 \textbf{Intent Acc.} & $m$ & es & de & zh & ja & pt & fr & hi & tr & \textbf{AVG} \\ 
 \hline
 English-Only Baseline* & 1 & 94.42  & 94.29  & 79.53  & 73.75  & 92.90 & 93.86  & 67.06  & 69.71  & 83.19  \\ 
  $Joint_{en-only}$ Baseline* & 1 & 95.03 & 94.51 & 80.54 & 73.57 & 93.48 & 93.33 & 73.53 & 71.05 & 84.38 \\
 Word-level CS & 1 & 94.18 & 93.92 & 81.67 & 75.48 & 92.54 & 94.18 & 81.19 & 74.22 & 85.92 \\
 Sentence-level CS & 1  & 94.60 & 93.53 & 81.21 & 75.01 & 93.10 & 93.24 & 82.37 & 75.11 & 86.02\\
 Chunk-level CS (CCS) & 1 & 95.12 & \textbf{95.27} & 83.88 & 74.27 & \textbf{94.20} & 93.48 & 82.73 & 77.51 & 87.06 \\
 $Joint_{en-only}$* + CCS & 1 & \textbf{95.48} & 94.51 & \textbf{84.43}$^\spadesuit$ & \textbf{76.48}$^\spadesuit$ & 94.15$^\spadesuit$ & \textbf{94.89}$^\spadesuit$  & \textbf{85.37}$^\spadesuit$  & \textbf{78.04}$^\spadesuit$ & \textbf{87.92} \\
\hline
 %Fully Supervised (Upper Bound) & 96.06 & 96.98 & 95.01 & 93.59 & 96.55 & 96.48 & 87.55 & 82.85 & 93.13 & 8 \\
 \multicolumn{11}{c}{Upper Bound} \\
 \hline
 Translate-Train (TT)* & 8 & 94.02 & 93.84 & 90.21 & 84.19 & 95.66 & 94.54 & 85.08 & 85.79 & 90.42 \\
 $Joint_{TT}$* & 8 & 94.16 & 94.24 & 91.56  & 85.98  & 95.75 & 95.01 & 86.45 & 84.95 & 91.01 \\
$Joint_{TT}$* + CCS & 8 & 95.48 & 95.41 & 91.60 & 87.17 & 95.34 & 94.60 & 87.94 & 85.93 & 91.68 \\
    \hline
    \hline
 \textbf{Slot F1} & $m$ & es & de & zh & ja & pt & fr & hi & tr & \textbf{AVG} \\ 
 \hline
 English-Only Baseline* & 1 & 96.16 & 96.73 & 83.12 & 78.81 & 95.63 & 95.40 & 77.05 & 88.09 & 88.87 \\
 $Joint_{en-only}$ Baseline* & 1 & 96.12 & 96.76 & 84.95 & 79.60 & 95.76 & 95.76 & 77.63 & 88.92 & 89.44 \\
 Word-level CS & 1 & 95.81 & 96.33 & 85.46 & 79.33 & 96.27 & 95.08 & 79.10 & 86.86 & 89.28 \\
 Sentence-level CS & 1  & 96.57 & \textbf{96.92} & 86.32 & 79.52 & \textbf{96.65} & 95.84 & 81.94 & 89.84 & 90.45 \\
 Chunk-level CS (CCS) & 1 & \textbf{96.68} & 96.82 & 87.10 & 80.00 & 96.46 & \textbf{96.31} & 80.95 & \textbf{91.60} & 90.51 \\
 $Joint_{en-only}$* + CCS & 1 & 96.09 & 96.56 & \textbf{88.61}$^\spadesuit$ & \textbf{82.28}$^\spadesuit$ & 96.01 & 95.94 & \textbf{82.28}$^\spadesuit$ & 90.45$^\spadesuit$ & \textbf{91.03} \\
\hline
 %Fully Supervised (Upper Bound) & 98.17 & 99.22 & 97.65 & 96.69 & 98.81 & 98.24 & 93.86 & 94.66 & 97.16 & 8 \\
 \multicolumn{11}{c}{Upper Bound} \\
 \hline
 Translate-Train (TT)* & 8 & 96.89 & 96.04 & 93.48 & 85.29 & 96.35 & 96.02 & 82.03 & 91.21 & 92.16 \\
  $Joint_{TT}$* & 8 & 96.92 & 95.66  & 93.64 & 87.84 & 96.11  & 95.95 & 82.98 & 91.15 & 92.53 \\
 $Joint_{TT}$* + CCS  & 8 & 96.98 & 96.27 & 93.37 & 85.87 & 95.88 & 95.44 & 82.00 & 91.31 & 92.14 \\
    \hline 
\end{tabular}
}
 \caption{Performance evaluation of code-switching with setting $k=5$. $CS$: Code-Switching. Reported scores are average of 5 independent runs (including a separate code-switched data for each run). $m$ = number of distinct models to be trained. *: modified BERT-based implementations \cite{chen2019bert,xu2020end}.\\ $^\spadesuit$: The difference is significant with p < 0.05 using Tukey HSD (conducted between $Joint_{en-only}$ + CCS versus $Joint_{en-only}$ Baseline for each language). } \label{table:style_score}
\end{table*}

\subsection{Joint Training}

\noindent Joint training is traditionally used for intent prediction and slot filling to exploit the correlation between the two tasks. This is done by feeding the feature vectors of one model to another or by sharing layers of a neural network followed by training the tasks together. So, a standard joint model loss can be defined as a combination of intent ($L_{i}$) and slot ($L_{sl}$) losses. i.e., $L = \alpha L_{i} + \beta L_{sl}$, where $\alpha$ and $\beta$ are corresponding task weights. Prior works \cite{goo2018slot,schuster2018cross,liu2016attention, niu2019novel} that use BiLSTM or RNN are now modified to BERT-based implementations explored in more recent works \cite{chen2019bert,hardalov2020enriched,xu2020end}. A standard $Joint$ model consists of BERT outputs from the final hidden state (classification (CLS) token for intent and $m$ word tokens for slots) fed to linear layers to get intent and slot predictions. Assuming $h_{cls}$ represents the CLS token and $h_{m}$ represents a token from the remaining word-level tokens, the BERT model outputs are defined as \cite{chen2019bert,xu2020end}:
\begin{align}
\label{eqn:1}
\begin{split}
p^{i}&=softmax(W^i h_{cls} + b^i)
\\
p^{sl}_m&=softmax(W^{sl} h_{m} + b^{sl}) \ \  \forall m
\end{split}
\end{align}
with a multi-class cross-entropy loss\footnote{ $L = -  \frac{1}{n} \sum_{i=1}^{n}  [y \log \hat{y}]$} for both intent ($L_i$) and slots ($L_{sl}$). We will use this model as our baseline for joint training. Our goal will be to show that code-switching on top of joint training improves the performance. The output of Algorithm \ref{alg:style} will be the input used for joint training on BERT for code-switched experiments.

\begin{table}
%\small
%\resizebox{\columnwidth}{!}{%
\centering
\scalebox{0.8}{
\begin{tabular}{lc}
    \hline
 \textbf{Intent Acc.} & ht  \\
 \hline
 English-Only Baseline* & 56.12  \\
 Chunk-level CS (CCS) & 63.15  \\ 
 $Joint_{en-only}$* + CCS  & \textbf{63.73}  \\
 \hline
  Translate-Train (TT)* & 62.58 \\ 
    \hline
    \hline
 \textbf{Slot F1} & ht \\
 \hline
 English-Only Baseline* & 68.72  \\
 Chunk-level CS (CCS)  & \textbf{70.27}  \\ 
 $Joint_{en-only}$* + CCS  & 70.02  \\
 \hline
 Translate-Train (TT)*  & 69.96  \\ 
    \hline 
\end{tabular}
}
 \caption{Performance on disaster data in Haitian Creole (ht). $CS$ = Code-Switching. Reported scores are average of 5 independent runs (*: modified BERT-based).} \label{table:crisis}
\end{table}

\section{Datasets}

\subsection{Benchmark Dataset}
\noindent We use the latest multilingual benchmark dataset of MultiATIS++ \cite{xu2020end}, which was created by manually translating the original ATIS \cite{price1990evaluation} dataset from English (en) to 8 other languages: Spanish (es), Portuguese (pt), German (de), French (fr), Chinese (zh), Japanese (ja), Hindi (hi), and Turkish (tr). The dataset consists of utterances for each language with an \textit{`intent'} label for \textit{`flight intent'} and \textit{`slot'} labels for the word tokens in BIO \cite{ramshaw1999text} format. A sample datapoint in English is shown in Figure \ref{fig:chunk}. 

\subsection{New Dataset for Disaster NLU}

\noindent We construct a new intent and slot filling dataset %by labelling 
of tweets collected during natural disasters, in two languages: English and Haitian Creole. The tweets originally were released by Appen\footnote{\url{https://appen.com/datasets/combined-disaster-response-data/}}. %From this, %we create a human-annotated dataset in two languages. 
For English, a language expert coded the tweets, and for Haitian Creole, we used Amazon Mechanical Turk with five annotators. %Here, intent prediction is as binary classification problem with two
Intent classes include: `\textit{request}' and `\textit{others}'. Slot filling consists of $5$ labels: `\textit{medical\_help}', `\textit{food}', `\textit{water}', `\textit{shelter}', and `\textit{other\_aid}'. Table \ref{table:stat} provides the dataset statistics. % for the tweet dataset. 

\begin{figure*} [ht]
    \centering
   \includegraphics[width=15cm,height=9cm]{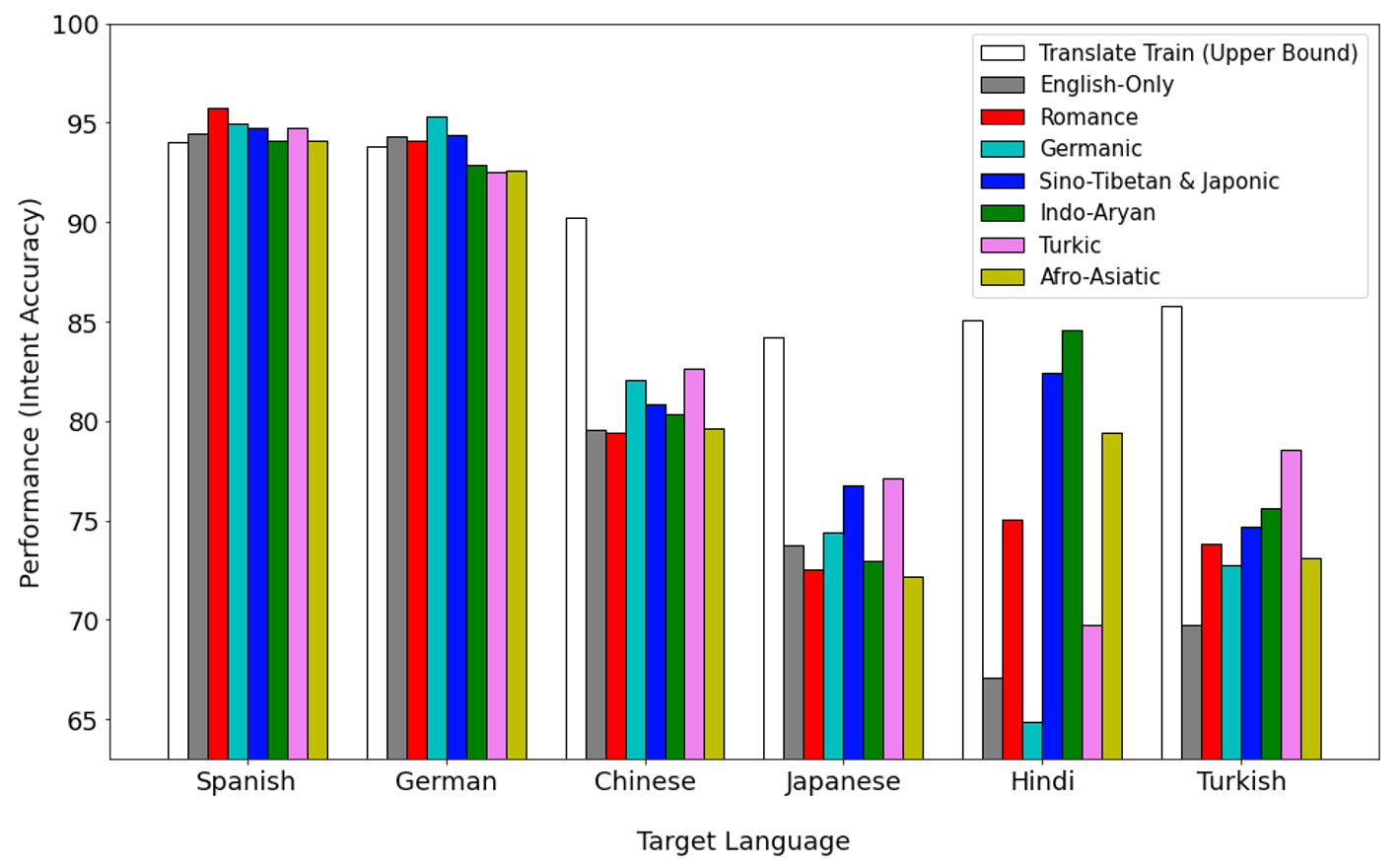}
    \caption{Impact of different language groups on the target languages.} \label{fig:group}
\end{figure*}

\section{Experimental Setup}

\noindent We use the traditional cross-lingual task setting where each experiment consists of a source language and a target language. A model is trained on the source data (English) and evaluated on the target data (8 other languages). For code-switching experiments, an English text is augmented with multilingual code-switching before training. Our  implementation is in PyTorch  \cite{NEURIPS2019_9015} and we use the pre-trained  \textit{bert-base-multilingual-uncased} \cite{devlin2018bert} with 
\textit{BertForSequenceClassification} \cite{wolf2020transformers} as the mBERT model. Maximum epoch is set to $25$ with an early stopping patience of $5$, batch size of $32$, and Adam optimizer \cite{kingma2014adam} with a learning rate of $5\mathrm{e}{-5}$. We select the best model on the validation set. Consistent with the metrics reported for intent prediction and slot filling evaluation in the past, we also use accuracy %to measure
for intent %performance 
and micro F1 to measure slot performance.

\subsection{Baselines \& Upper Bound}
\noindent Since we assume that target language is not known before hand, \textbf{Translate-Train (TT)} \cite{xu2020end} method is not a suitable baseline. Rather, we set this to be an upper bound, i.e. translating to the target language and fine-tuning the model should intuitively outperform a generic model. Additionally, we add code-switching to this TT model to assess if augmentation negatively impacts its performance. The zero-shot baselines for the code-switching experiments use an \textbf{English-Only} \cite{xu2020end} model, which is fine-tuned over the pre-trained mBERT separately for each task and an English-only \textbf{Joint} model \cite{chen2019bert}.

\begin{table} %[tbp!]
%\small
%\resizebox{\columnwidth}{!}{%
\centering
\scalebox{0.85}{
\begin{tabular}{ccccc}
    \hline
 \textbf{CS (k=5)} & \textbf{MTT} & $\mathbf{Joint_{en}}$ & $\mathbf{Joint_{cs}}$ & $\mathbf{Joint_{TT}}$  \\
 \hline
 05:04:49 & 1:31:32 & 00:11:50 & 01:06:50 & 00:11:04   \\
    \hline
\end{tabular}
}
%\caption[Caption for LOF]{\footnotemark hello}
 \caption{Runtime on Google Colab (K80 GPU for training joint models). $MTT$: Machine Translation to Target. Note that $MTT$ and $J_{TT}$ are for one target language (averaged).}  \label{table:runtime}
\end{table}

%\textit{es}, \textit{zh}, and \textit{hi}
%\footnotemark
%\footnotetext{$\alpha=\beta=1.0$, $epochs=25$, $lr=5\mathrm{e}{-5}$, $freeze<=4$}

%is for one target language and it combines monolingual source with its target-translated data.

\section{Results \& Discussion}

\subsection{Effect of Multilingual Code-Switching}
\noindent Table \ref{table:style_score} describes performance evaluation on the MultiATIS++ dataset. When compared to the state-of-the-art jointly trained English-only baseline, we see a $+4.2\%$ boost in intent accuracy and $+1.8\%$ boost in slot F1 scores on average by augmenting the dataset via multilingual code-switching \textbf{without requiring the target language}. From the significance tests, except for Spanish and German, all other languages were helped by code-switching for intent detection. For slot filling, improvement on Portuguese and French went insignificant. This suggests that code-switching primarily helped languages that are morphologically more different as compared to the source language (English). For example, Hindi and Turkish have the highest intent performance improvement of $+16.1\%$ and $+9.8\%$ respectively. And for slots, Hindi and Chinese with $+6.0\%$ and $+4.3\%$ respectively. Japanese showed $+4\%$ improvement for intent and $+3.4\%$ for slots.

The running time of the models in Table \ref{table:runtime} show that code-switching is expensive which can take up to $5$ hours for $k=5$. Its training is also expensive because there is $k$ times more data as compared to the monolingual source data. Increasing the number of code-switchings ($k$) for a sentence from $5$ to $50$ improved the performance by $+1\%$, while increasing the run-time by a large margin. So, parameter $k$ should be picked appropriately. Albeit this time cost is for training, with benefits at the prediction stage for real world problems. %, being cautioned while adding a lot of code-switched data. 

%; when compared to the translate-train approach which requires 8 models to be trained during test time. 

%On the other hand, i
In the translate-train (upper bound) scenario, it is not immediately clear if augmentation can help, because data in the same language as the target is always preferred over other languages, or code-switched. However, we show in Table \ref{table:style_score} that augmentation did not hinder the performance.

For both intent and slot performance, chunk-level model  remained robust across the languages. For intent, difference between word-level and sentence-level was insignificant. For slot, sentence-level was in par with chunk-level on average. Thus, we think that code-switching at chunk-level is safer for avoiding semantic discrepancies (as in the word-level) while also capturing better intra-sentence language neutrality.

\begin{figure} [t]
    \centering
    \includegraphics[scale=0.22]{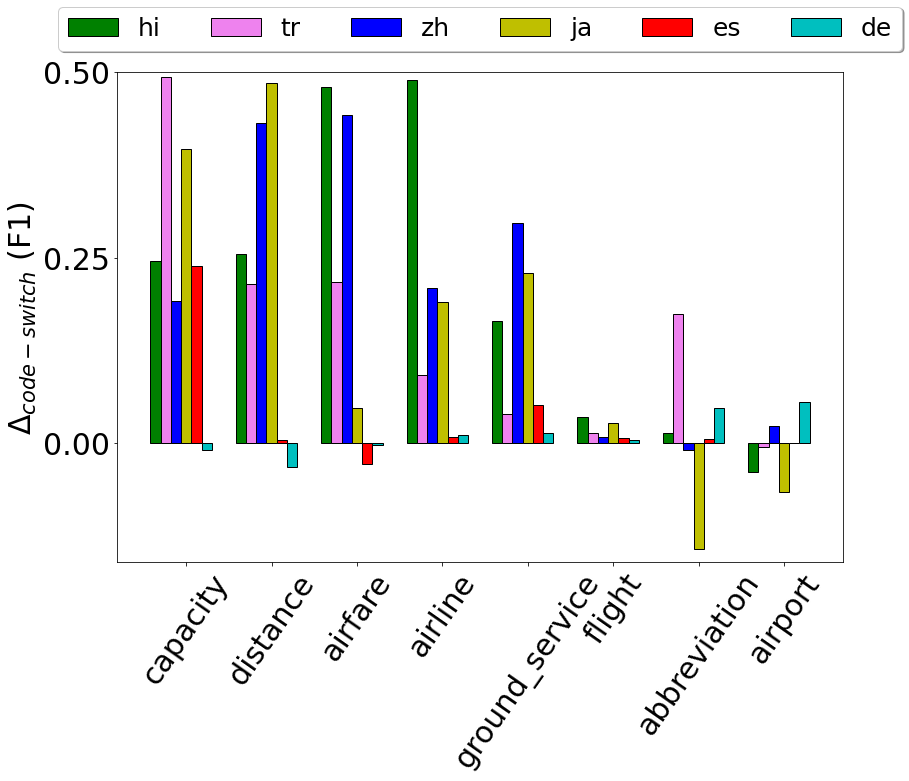} %[width=15cm, height=8cm]
    \caption{Impact of code-switching on intent classes.} \label{fig:hi_intent}
\end{figure}

\begin{figure} [t]
    \centering
    \includegraphics[scale=0.21]{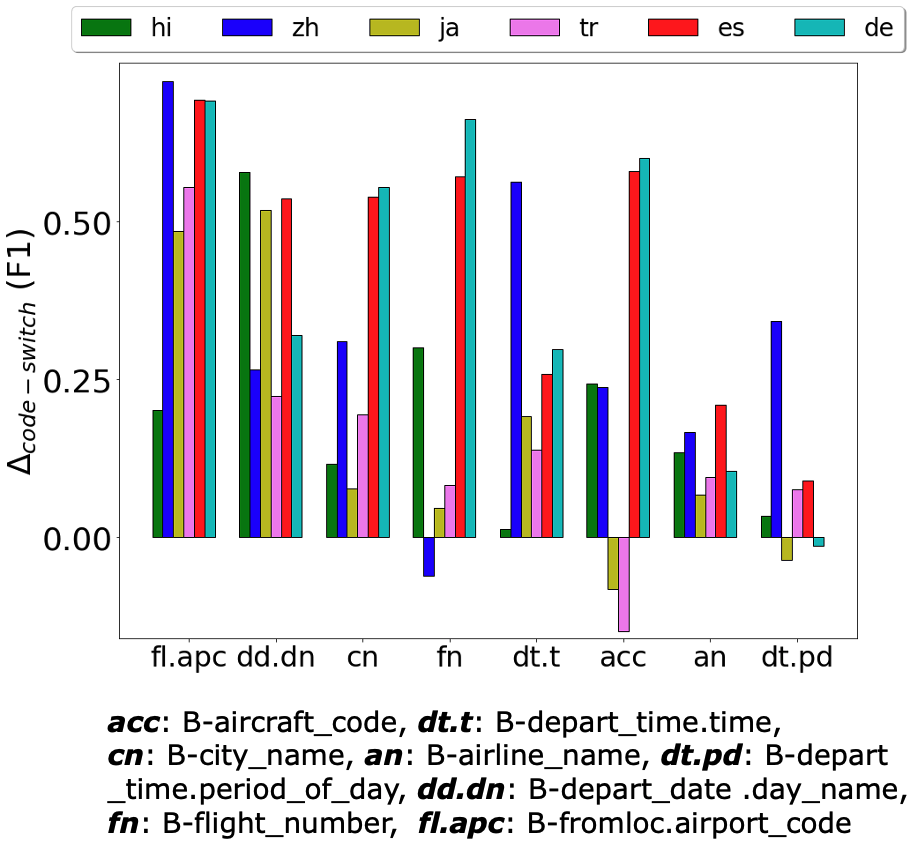} %[width=15cm, height=8cm]
    \caption{Impact of code-switching on slot labels.} \label{fig:ja_slot}
\end{figure}

\begin{figure} [t]
    \centering
    \includegraphics[scale=0.38]{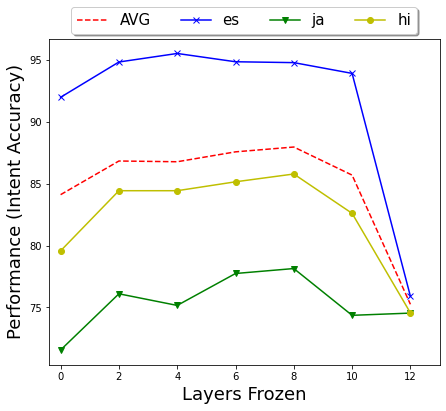} %[width=15cm, height=8cm]
    \caption{Freezing earlier layers and unfreezing a few at the top of the transformer appear to be most optimal.} \label{fig:freeze}
\end{figure}

\subsection{Evaluation on Disaster Dataset }%Haitian Creole}
\noindent We found that disaster data is more challenging when compared to the ATIS dataset for transfer learning in NLU. The predictive performance is shown in Table \ref{table:crisis}. Code-Switching improved intent accuracy by $+12.5\%$ and slot F1 by $+2.3\%$, which is promising considering that they are tweets. Joint training added $+0.9\%$ improvement to intent accuracy, however did not seem to help slot F1. This might imply a lack of strong correlation between the two tasks, i.e. a mention of `\textit{food}' or `\textit{shelter}' in a tweet may not always mean that it is a `\textit{request}' or vice-versa. The upper bound of translate-train method did not perform any better than the randomly code-switched model which seemed  counter-intuitive. %We speculate that t
This might be due to the lack of strong representation for Haitian Creole in the pre-trained model, although it is similar to French.

\subsection{Impact of Language Families}
\noindent Results of language family analysis are shown in Figure \ref{fig:group}. The 
%original dataset 
input 
in English is independently code-switched using 6 different language families. Note that the target language is always excluded from the group when evaluating on the same, i.e. Hindi is excluded from Indo-Aryan family when that family is being evaluated on it. Translate-train model is provided as a frame of reference and upper bound. We dropped French and Portuguese from the chart as they fall in to Romance family similar to Spanish. %From the results, we see 
Results show the language families helped their corresponding languages, i.e. Romance helped Spanish, Germanic helped German, %Indo-Aryan helped Hindi, 
and so on; with the exception of Chinese and Japanese. In both cases, Turkic language family helped %stood out 
better than others. %, which is intriguing. 
%This encourages a further inquiry %linguistic investigation into once rejected hypothetical language family of Altaic \cite{georg1999telling} to study the relation between Turkic family and Japanese. 

% as shown in Figure \ref{fig:hi_intent},
\subsection{Error Analysis}
\noindent Selecting intent classes with support $>10$, Figure \ref{fig:hi_intent} shows how each class is positively or negatively impacted by code-switching. Improvement was primarily on `\textit{airfare}', `\textit{distance}' `\textit{capacity}', `\textit{airline}', and `\textit{ground\_service}' which had longer sentences such as `\textit{Please tell me which airline has the most departures from Atlanta}' when compared to `\textit{abbreviations}' and `\textit{airport}' classes that included very short phrases like `\textit{What does EA mean?}' However, note that, Spanish and German did not improve much; aligning with our results in Table \ref{table:style_score}. For slot labels in Figure \ref{fig:ja_slot}, we selected the ones with support $>50$ and that have different characteristics, e.g.  `\textit{name}', `\textit{code}', etc. %The plot reflects 
The overall trend in slot performance shows improvements for labels such as `\textit{day\_name}', `\textit{airport\_code}', and `\textit{city\_name}' and slight variations in labels such as  `\textit{fight\_number}' and `\textit{period\_of\_day}'; implying  textual slots benefiting over numeric ones.

%an intuitive improvement was found for text labels such as `\textit{day\_name}' and `\textit{month\_name}' when compared to labels such as `\textit{airline\_code}', `\textit{day\_number}', and `\textit{year}'. However, there were also counter-intuitive labels such as `\textit{restriction\_code}', `\textit{fare\_amount}', and `\textit{time}' that was improved by code-switching.

\subsection{Hyperparameter Tuning}
\noindent %In addition to 
For joint training with same task weights, we tuned $\alpha$ and $\beta$ using grid search to see the strength of correlation between the tasks. For intent, the ($\alpha,\beta$) combination of ($1.0,0.6$) performed well, while ($1.0,1.0$) for slots. This suggests that intent benefiting slot might be slightly more than slot benefiting intent. Additionally, during fine-tuning, freezing the layers of the transformer affected the model performance as shown in Figure \ref{fig:freeze}. Keeping the first $8$ layers frozen gave the best performance. By freezing the earlier layers, the transformer can retain its most fundamental feature information gained from the massive pre-training step, and by unfreezing some top layers, it can undergo fine-tuning.

%We ran into a token disparity problem in BERT model training where the performance is impacted when there is big difference between the token lengths of train and test data; particularly for the tweet dataset. We leave a further investigation into this athe masking as future work.

\section{Conclusion \& Future Work} 

\noindent This study shows that augmenting the monolingual input data with multilingual code-switching via random translations helps a zero-shot model to be more language neutral when evaluated on unseen languages. % in a pure  setting. 
This approach enhanced the generalizability of pre-trained mBERT when fine-tuning for downstream tasks of intent detection and slot filling. We presented an application of this method using a new annotated dataset of disaster tweets. Further, we studied code-switching with language families and their impact on specific target languages, which can be used to enhance the zero-shot generalizability of models created for low-resource languages. Expanding to XLM-R and similar approaches to improve masked language model training by addressing code-switching during pre-training and releasing a larger dataset of annotated disaster tweets in more languages are planned for future work. \\ 
%\noindent{\textbf{Reproducibility}:} We provide source code and datasets as supplementary material. 

\section{Acknowledgement}
We thank U.S. National Science Foundation grants IIS-1815459 and IIS-1657379 for partially supporting this research. We also thank Ming Sun and Alexis Conneau for giving valuable insights on multilingual model training. We also acknowledge ARGO team as the experiments were run on ARGO, a research computing cluster provided by the Office of Research Computing at George Mason University.

% Entries for the entire Anthology, followed by custom entries
\bibliography{anthology,custom} %anthology,
\bibliographystyle{acl_natbib}

\end{document}